# ENGLISH TO ARABIC MACHINE TRANSLATION OF MATHEMATICAL DOCUMENTS

Mustapha Eddahibi and Mohammed Mensouri

IMIS Laboratory, Ibnou Zohr University – Agadir Morocco

## ABSTRACT

*This paper is about the development of a machine translation system tailored specifically for LATEX mathematical documents. The system focuses on translating English LATEX mathematical documents into Arabic LATEX, catering to the growing demand for multilingual accessibility in scientific and mathematical literature. With the vast proliferation of LATEX mathematical documents the need for an efficient and accurate translation system has become increasingly essential. This paper addresses the necessity for a robust translation tool that enables seamless communication and comprehension of complex mathematical content across language barriers. The proposed system leverages a Transformer model as the core of the translation system, ensuring enhanced accuracy and fluency in the translated Arabic LATEX documents. Furthermore, the integration of RyDArab, an Arabic mathematical TEX extension, along with a rule-based translator for Arabic mathematical expressions, contributes to the precise rendering of complex mathematical symbols and equations in the translated output. The paper discusses the architecture, methodology, of the developed system, highlighting its efficacy in bridging the language gap in the domain of mathematical documentation.*

## 1. INTRODUCTION

Machine Translation (MT), is the conversion of text from one language to another using computers. This process involves the utilization of programming languages and software to accomplish the translation tasks [1]. Various types of content, ranging from commercial and business documents to scientific papers, instruction manuals, textbooks, and online materials, require translation services. Presently, machine translation has found widespread application in the language services industry and has become a common tool for professional translators. However, machine translation has limitations and is not suitable for every type of textual content. The challenges in Machine Translation stem from the disparities in grammatical structures existing between the source and target languages. Furthermore, the complexity of translation can be influenced by the specific category of the text. While the translation of everyday spoken language is relatively straightforward, rendering poetry, philosophical treatises, scientific or technical documents poses more intricate challenges [2]. The seamless translation of complex mathematical documents from English to Arabic stands as a big challenge in the machine translation field. Significant progress has been achieved in this domain, the intricacies of mathematical language, particularly in the context of LATEX [3] syntax, present a critical gap that demands attention.

This study delves into uncharted territory, aiming to bridge the gap in the translation of mathematical content into the Arabic language, a domain remains relatively unexplored in the current body of literature. Leveraging cutting-edge methodologies such as LATEX parsing, mathematical expression tokenization, and the utilization of a Transformer for natural language translation, alongside a rule-based translator for managing mathematical expressions, our research aspires to contribute significantly to the burgeoning field of automated translation.

By building upon recent advancements in pre-trained language models, we seek to unravel the complex interplay between mathematical expressions and linguistic structures, thereby unlocking the potential for comprehensive and accurate translation.

Despite the acknowledging the inherent intricacies linked with handling symbolic mathematical content, the absence of a suitable corpus remains a notable limitation. The intricate nature of mathematical language, coupled with the unique challenges posed by Arabic notation, necessitates a meticulous and nuanced approach to the translation process.

This research holds substantial promise for the educational landscape, benefitting researchers, educators, and students alike. By facilitating a deeper comprehension of intricate mathematical documents, our work endeavours to streamline the creation, modification, and dissemination of these materials in both English and Arabic contexts. The digitized representation of such documents not only fosters their creation and archiving but also enables their seamless exchange via various communication networks. However, realizing this objective necessitates overcoming several challenges, particularly in dealing with the distinctive nature of mathematical expressions in Arabic notation.

## 2. RELATED WORK

The prominence of English as the universal language of science emerged a mere four centuries ago, marking a significant turning point in global communication. Its unparalleled expansion across the world post-World War II solidified its dominance, surpassing the influence of any other language in history [4]. This trend has underscored the democratization of scientific knowledge, granting access to a wider audience beyond traditional linguistic barriers. Nonetheless, it emphasizes the critical importance of promoting education in one's native language, highlighting the necessity for striking a balance between the advantages of a common scientific language and the preservation of cultural and linguistic diversity for comprehensive learning.

The recognition of the significance of automated translation emerged many years back. Numerous studies have focused on machine translation within the scientific realm. Tehseen et al. introduced an English-to-Urdu scientific text translator that employs term tagging and domain-specific translation [5]. This translator was specifically designed for the translation of computer science documents and was assessed using a self-generated corpus within the computer science domain. The translator algorithm didn't show mathematical expressions translation or used file format.

Because of the widespread adoption of LATEX stems owing to its capacity to manage complex mathematical expressions, equations, and symbols with unparalleled precision and elegance, the translation of documents in such format is a great opportunity. Ohri et al. developed a machine translation system named Polymath for converting LATEX documents with mathematical text [6]. This system is capable of translating English LATEX content to French LATEX. It operates by transforming the main content of an input LATEX document into English sentences with mathematical tokens, followed by the application of a pre-trained Transformer-based translator model.

The process of translating LATEX documents from English to French is relatively straightforward compared to the challenges encountered when translating from English to Arabic. This discrepancy arises from the fundamental differences between the Arabic and English mathematical notations. The Arabic mathematical system, distinct in its structure and representation, introduces complexities that are not encountered in the French translation process. The nuances of Arabic mathematical symbols and expressions often require specific handling techniques and an in-depth understanding of the language's unique grammatical rules and conventions. Despite the potential complexities, the English to French LATEX translation process proves to be more seamless due to the greater similarity in notation and linguistic constructs between the two languages.

## 3. ARABIC MATHEMATICAL NOTATION

In Arabic scientific documents there are two models of mathematical expressions: mixed and Pure Arabic notations. Mixed mathematical notation used in some countries like Morocco, is the outcome of a literal word-to-word translation of mathematical French books. The syntactic models of several formulations have been transported to the new language. The symbolic writing was imported just as it was, without any changes.

$$f(x) = \begin{cases} \sum_{i=1}^{s} x^i & \text{إذا كان} \quad x < 0 \\ \int_{1}^{s} x^i \, dx & \text{إذا كان} \quad x \in E \\ \text{tg } \pi & \text{غير ذلك} \ (\text{مع} \ \pi \simeq 3,141) \end{cases}$$

Figure 1. Mixed Mathematical Notation

In the pure Arabic presentation, mathematical expressions spread from right to left and use Arabic symbols from its alphabet. These symbols are used to note unknown variables and functions names. As for common functions that are replaced by their abbreviated names. In this notation we can distinguish two types. The first uses some mirrored latin symbols like the sum ⟨mirrored sum symbol⟩. The second type uses some Arabic alphabet-based symbols like ⟨Arabic sum symbol⟩ for the sum.

Figure 2. Pure Arabic Notation with mirrored symbols

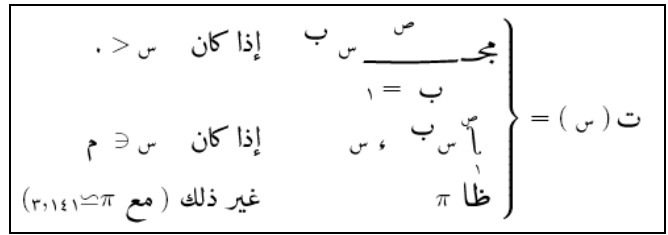

Figure 3. Pure Arabic Notation with Arabic alphabet-based symbols

## 4. TRANSLATION OF THE LATEX DOCUMENT

The process of translating the LATEX document comprises two distinct phases, each contributing to the accurate conversion of its contents. Initially, the document undergoes a comprehensive parsing phase, allowing for the precise identification and isolation of its constituent elements. Subsequently, the second phase entails the systematic extraction of mathematical expressions, subsequently replaced by an indexed list of tokens (Exprr1, Exprr2, etc.). These identified expressions, denoted as Exprri, serve as novel entities during the natural language translation process, functioning as placeholders for the mathematical components. Consequently, the mathematical expressions are stored within a list, with the corresponding tokens acting as their respective indices. The subsequent translation process relies on a rule-based translation function tailored to the unique characteristics of mathematical notations. This function operates using the specific mathematical expression, its associated token, and the corresponding notation type, ensuring the seamless and accurate transformation of the mathematical content within the document.

### 4.1. English to Arabic Natural Language Translator

The natural language translation part of the system is based on transformer architecture [7]. Opting for transformers over RNNs is supported by their capacity to process input data in parallel. They are not susceptible to the vanishing gradient issue. Transformers proficiently grasp intricate connections among various components of the input and utilize attention mechanisms to focus on relevant portions of the input sequence [8].

Before the translation a pre-processing phase is done. The content of parsed LATEX blocs is segmented to sentences. Certain formatting commands, such as \textbf and \textit, which alter the style of individual or multiple words within a sentence, are eliminated to streamline the text. This process aims to enhance the translation, particularly considering that Arabic typographic conventions do not employ bold, italic, or roman styles, among others.

The transformer undergoes training and validation using datasets comprising English words, sentences, and paragraphs along with their corresponding Arabic translations. Its architecture consists of two primary components: an encoder and a decoder for sequence-to-sequence translation, alongside an attention mechanism. Figure 4 provides an overview of the system.

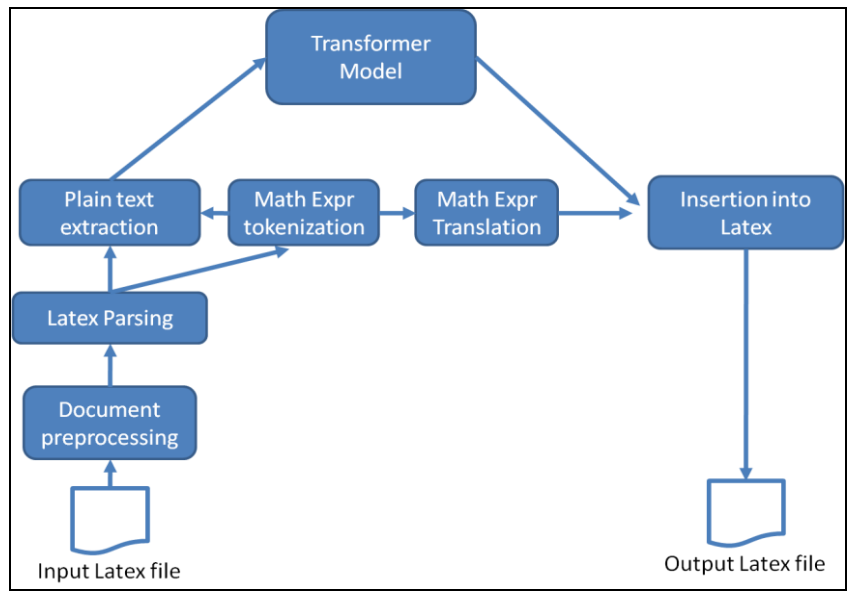

Figure 4. Overview of the developed system

## 4.2. Translation of Mathematical Expression to Arabic Notation

The authors note that the existing pool of translations from English to Arabic for mathematical expressions lacks sufficiency for the utilization of neural network-based translation methods. Consequently, the current optimal approach is rule-based translation, despite its non-exhaustive nature [9]. The translation process involves the following steps:

- ♦ Utilization of TEX[10] packages, including RyDArab[11] for typesetting Arabic mathematical expressions and curext [12] for handling stretched mathematical symbols in the resulting LATEX file.
- ♦ Introduction of the command \arabmath to each mathematical expression to ensure proper alignment.
- ♦ Transliteration of alphabetical symbols into their Arabic equivalents.
- ♦ Implementation of the commands \warabnum for standard Western Arabic digits 9،8،7،6،5،4،3،2،1،0 or \earabnum for Eastern Arabic digits (Hindi digits) ٩،٨،٧،٦،٥،٤،٣،٢،١،٠.
- ♦ Use of \alp without dots for alphabetic symbols without dots ( ح ) and \alpwithdots for those with dots ( ج ).
- ♦ Application of \fun with dots and \fun without dots commands for elementary functions with and without dots, respectively ( جا, حا ).

When these options are incorporated into the preamble of the input file, the commands are universally applied to all mathematical expressions. To address document encoding concerns, the output file must adhere to either the ISO-8859-6 encoding system for ArabTEX [13] or UTF-8 for Omega [14].

♦ Furthermore, specific transformations are applied to mathematical symbols, such as replacing \sum with \lsum ( ), \csum ( ) for its Arabic literal linear and curved equivalents, and \ssum ( ) for the mirrored equivalent. Similar transformations are implemented for the product command.

## 5. RESULTS, LIMITATIONS AND PERSPECTIVES

For system evaluation, we conducted a meticulous assessment using targeted documents that explore mathematical concepts similar to the ones discovered in textbooks used in primary and secondary education courses. The results for translating mathematical expressions were generally satisfactory, showcasing the system's competency. However, some discrepancies were observed, which can be attributed to the inherent limitations of the non-exhaustive rule-based method.

Notably, the machine translation performance, utilizing the transformer model, exhibited markedly low BLUE scores. The BLEU (Bilingual Evaluation Understudy) score [15] is a frequently employed measure for assessing the quality of machine-generated translations by comparing them to human-generated references. In this context, the lower BLUE scores indicate a misalignment between the machine-generated translations and the expected human references. This suboptimal performance can be traced back to two key factors: the relatively modest size of the dataset used for training and the substitution of mathematical expressions with "exprri" tokens in the natural language input.

The limited dataset size may have hindered the model's exposure to diverse linguistic patterns, leading to challenges in accurately capturing the nuances of natural language. Additionally, the abstraction introduced by the "exprri" tokens may have contributed to a loss of contextual information during the translation process, further impacting the system's ability to generate linguistically fluent and contextually accurate translations. Addressing these issues by augmenting the dataset and refining the tokenization approach could potentially enhance the system's overall translation performance.

In future work, we aim to advance towards a comprehensive machine translation system capable of translating entire LATEX documents, including intricate LATEX commands, into Arabic. This expansion is driven by the integration of the DadTEX [16] extension, designed for completely Arabic TEX environments. This ambitious undertaking holds potential to enhance the accessibility of scientific literature by providing seamless translation of diverse document components. The incorporation of DadTEX signifies a crucial step towards facilitating cross-linguistic communication in scientific and mathematical domains. The future perspectives emphasize a commitment to overcoming language barriers and promoting inclusivity in scientific discourse.

## 6. CONCLUSIONS

In conclusion, this paper introduces a specialized machine translation system for LATEX mathematical documents, concentrating on translating English LATEX mathematical content into Arabic LATEX. Driven by the escalating demand for multilingual accessibility in scientific literature, especially in the realm of complex mathematical expressions documented in LATEX, the proposed system employs a text-to-text transfer transformer to enhance accuracy and fluency in the translated Arabic LATEX documents.

The integration of RyDArab, an Arabic mathematical TEX extension, and a rule-based translator for Arabic mathematical expressions ensures precise rendering of intricate symbols and equations in the translated output. The discussed architecture, methodology, and performance evaluation underscore the system's effectiveness in overcoming language barriers in mathematical documentation.

Acknowledging the complexities of handling symbolic mathematical content, the study underscores the necessity for an adequate corpus to improve translation accuracy. Despite challenges posed by the unique nature of Arabic mathematical expressions, this research promises significant benefits for education, aiding researchers, educators, and students in comprehending intricate mathematical documents in both English and Arabic.

Addressing the unexplored realm of translating mathematical content into Arabic, this paper recognizes and navigates the specific challenges posed by Arabic mathematical notations. It also contributes valuable insights into related work, emphasizing the importance of automated translation in scientific communication, particularly focusing on LATEX documents. In summary, this study seeks to address a crucial void in automated translation, particularly in mathematical documents, laying the groundwork for further advancements in multilingual accessibility within the scientific and mathematical community.

**AUTHORS**

**Prof. Mustapha Eddahibi** received his PhD from Unversity Cadi Ayyad Marrakech in 2007. He is currently computer science teacher researcher in University Ibn Zohr. He is a former head of the decisional expert systems research team. His research interests are in the area of intelligent computing, information engineering, digital Information Encoding and Processing.

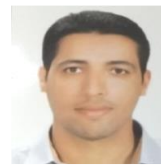

**Prof. Mohammed Mensouri** received the M.S degree in Networks and Telecommunication in 2008, from Faculty of Sciences and Technology, Cadi Ayyad University, Marrakech, Morocco. In 2015, He received Ph.D of Computer Science in Faculty of Sciences, Chouaib Doukkali University, El Jadida, Morocco. He is professor at Ibn Zohr University, Agadir, Moroc. His research interest information theory and channel coding , especially error correction codes.

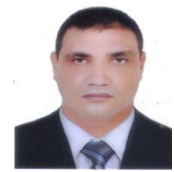